\documentclass[review]{elsarticle}
\usepackage{hyperref}
\usepackage{cite}
\usepackage{amsmath,amssymb,amsfonts}
\usepackage{algorithmic}
\usepackage{graphicx}
\usepackage{textcomp}
\usepackage{caption}
\usepackage{subcaption}
\usepackage{multirow}
\usepackage{xcolor}
\usepackage{listings}

\journal{Science of Computer Programming}

\biboptions{authoryear}

\begin{document}

\begin{frontmatter}

\title{Web-based Application for Detecting Indonesian Clickbait Headlines using IndoBERT}


\author{Muhammad Noor Fakhruzzaman\corref{mycorrespondingauthor}}
\ead{ruzza@stmm.unair.ac.id}

\author{Sie Wildan Gunawan}
\ead{sie.wildan.gunawan-2020@stmm.unair.ac.id}


\cortext[mycorrespondingauthor]{Corresponding author (tel. +62811868567)}

\address{Faculty of Advanced Technology and Multidiscipline, Universitas Airlangga, Indonesia}

\begin{abstract}
With increasing usage of clickbaits in Indonesian Online News, newsworthy articles sometimes get buried among clickbaity news. A reliable and lightweight tool is needed to detect such clickbaits on-the-go. Leveraging state-of-the-art natural language processing model BERT, a RESTful API based application is developed. This study offloaded the computing resources needed to train the model on the cloud server, while the client-side application only need to send a request to the API and the cloud server will handle the rest. This study proposed the design and developed a web-based application to detect clickbait in Indonesian using IndoBERT as a language model. The application usage is discussed and available for public use with a performance of mean ROC-AUC of 89\%.
\end{abstract}


\end{frontmatter}


\section{Introduction}
Clickbait headlines have becoming more prominent since the advertiser decide to count popularity based on clicks. Some news that were deemed as less newsworthy used clickbait to attract readers, to ensure that the news still makes money \citep{chen_news_2015}. Since then, a lot of scientific articles proposed ways to detect clickbait using Artificial Intelligence, one of them was fine tuned for Indonesian headlines \citep{chakraborty_stop_2016,chen_news_2015,anand_we_2017,agrawal_clickbait_2016,biyani__2016,maulidi_penerapan_2018,fakhruzzaman_clickbait_2021}. However, those articles often propose methods only but did not create any usable tool for the readers.

Some Indonesian people simply cannot distinguish clickbait and normal headlines and often get disappointed after clicking the news, and it may be correlated to the inequality of internet access in Indonesia, widely known as the digital divide \citep{puspitasari2016digital}. People with less internet exposure may have a hard time at pointing out clickbaits, simply because they seldom encountered it.

By the availability of an accessible tool, digital literacy among readers may increased and misinformation can be hindered \citep{hurst_clickbait_2016,chen_misleading_2015}. Furthermore, many methods in detecting clickbait from scientific articles can be implemented with high level of abstraction, so that people can use it easily without hassle.

Therefore, this study aims to propose and develop an easy-to-use application to detect Indonesian Clickbait Headline, leveraging state-of-the-art language model, while still being easy on the client's computing resource.

\section{Material and Methods}

This study uses an existing model and annotated dataset of clickbait headlines. The model architecture used Multilingual BERT and topped with a hidden layer of 100 neurons and one output layer \citep{fakhruzzaman_clickbait_2021}. While the dataset consisted of 6000 annotated headlines with balanced class of clickbait and non-clickbait, also with absolute reliability \citep{william_click-id_2020}.

\begin{figure}[!htb]
\center{\includegraphics[width=.9\columnwidth]
{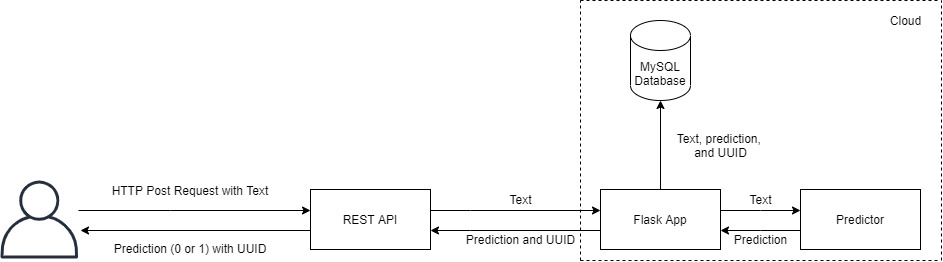}}
\caption{\label{flowpredict}User Request Flow Diagram}
\end{figure}

Figure \ref{flowpredict} shows the application architecture, depicting how user requests flow through the application. It leverages the REST API architecture.
REST API architecture offers flexibility that it can run on multiple front-end versions at the same time without breaking its clients \citep{li_design_2016}. Using HTTP request, it leverages the light protocol to send standardized data format, often in JSON, on a secure way because it still travels through HTTPS. This way, when handling requests from mobile applications, or from any client-side interface, the data still flows seamlessly to the cloud server.

In order to reduce memory usage, we train the model using IndoBERT Lite Base by IndoNLU instead of using Multilingual BERT model used in the previous study \citep{fakhruzzaman_clickbait_2021}. This pre-trained model has 12 layers and around 11.7 million parameters. Compared to other pre-trained models, IndoBERT Lite Base performs well with f1-score average of 85\%. Other models in IndoBERT family performs better but they require more parameters, hence higher memory usage \citep{wilie_indonlu_2020}.

\begin{figure}[!htb]
\center{\includegraphics[width=.9\columnwidth]
{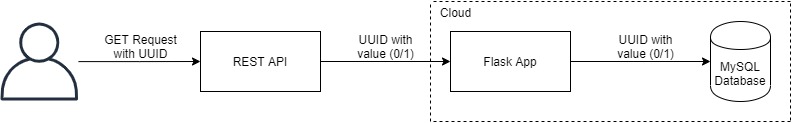}}
\caption{\label{dfdsql}User Feedback Flow Diagram}
\end{figure}

This study also use mysql as the database engine due to its compatibility with a lot of server-side environments \citep{kofler2001mysql}. The database is used to store user feedback of the prediction for later retraining purposes. Figure \ref{dfdsql} shows the feedback flow diagram and how the prediction is stored into the database.

Furthermore, this study use flask to build the API, fronted by a gunicorn WSGI and nginx webserver to serve the endpoint through the cloud, as depicted in Figure \ref{flowpredict}. Flask is used because it is not only simple and straightforward, but also very flexible, it provides high level abstraction to build API easily. Additionally, flask integrates seamlessly with gunicorn and nginx \citep{ghimire2020comparative}. Moreover, nginx is used for its reverse-proxy capability, so that in the future, other microservices can be added to the same server and contained easily. 

REST API (or sometimes called RESTful API) is an API that uses REST architecture to handle a request sent from a user in the front-end. This allows users to interact with RESTful web services. In short, REST API works like a mediator between user and server \citep{masse2011rest}.


The API was built to communicate the front-end with a trained model, while the training iteration was executed on a separate Google Colaboratory platform. A snippet of the API script is shown on the following script.

\begin{lstlisting}
--app.py--

from flask import Flask, request, jsonify, make_response
from flask_limiter import Limiter
from flask_limiter.util import get_remote_address
from predict import predict

@app.route("/predict", methods=["POST"])
@limiter.limit("2 per minute")
def predictText():
	request_data = request.get_json()
	uuid = uuid4()
	text = request_data['text']
	user_ip = request.remote_addr
	prediction = predict(text)
	data = UserRequest(
		uuid = uuid,
		text = text,
		prediction = prediction,
		ip_address = user_ip
	)
	db.session.add(data)
	db.session.commit()
	response = make_response(
		jsonify(
			{
				"id": uuid,
				"prediction": prediction
			}
		),
		200
	)
	response.headers["Content-Type"]="application/json"
	return response
\end{lstlisting}

The script is then hosted on a DigitalOcean droplet. Then, a static page hosted on github is used to invoke the API via HTTPS. The complete script is available in our github repository.

To secure the endpoint from any unwanted actions, this study used encrypted HTTPS protocol to serve both the API and the front-end. We also use a request limiter in the front-end and back-end to prevent spamming and DoS attacks. 

The limiter limits requests to one request per minute from the fronted but set to be two requests per minute in the back-end. The limit difference is due to Axios' default behaviour to send pre-flight request before sending the actual request.

\section{Results}
After training the model, IndoBERT performs well with an ROC-AUC average of 89\%. 

Using this approach, memory usage is reduced by around 800MB. Less memory usage means our model is more efficient and will use less resources when processing a prediction request. Therefore, the cost of the server that hosts the model can be significantly reduced.

\begin{figure}[!ht]
  \begin{subfigure}[b]{.5\columnwidth}
    
    \center{\includegraphics[width=\columnwidth]{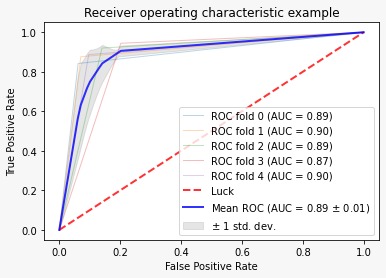}}
    
    \subcaption{\label{rocaucbaru}IndoBERT Lite Base ROC-AUC Plot}
    
  \end{subfigure}%
  \begin{subfigure}[b]{.5\columnwidth}
    \center{\includegraphics[width=\columnwidth]{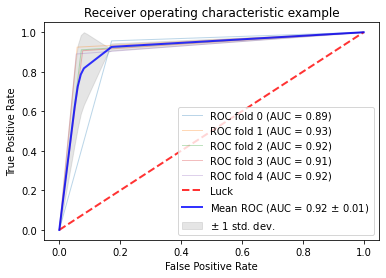}}
    
    \subcaption{Multilingual BERT ROC-AUC Plot}
    \label{rocauclama}
  \end{subfigure}
  \caption{Language Model Performance Comparison}
  \label{rocaucindobert}
\end{figure}

However, switching to lighter model sacrificed prediction performance. Figure \ref{rocaucindobert} shows the ROC-AUC plot of IndoBERT model used in the developed application compared to Multilingual BERT model used in previous study \citep{fakhruzzaman_clickbait_2021}.

\begin{figure}[!htb]
\center{\includegraphics[width=.5\columnwidth]
{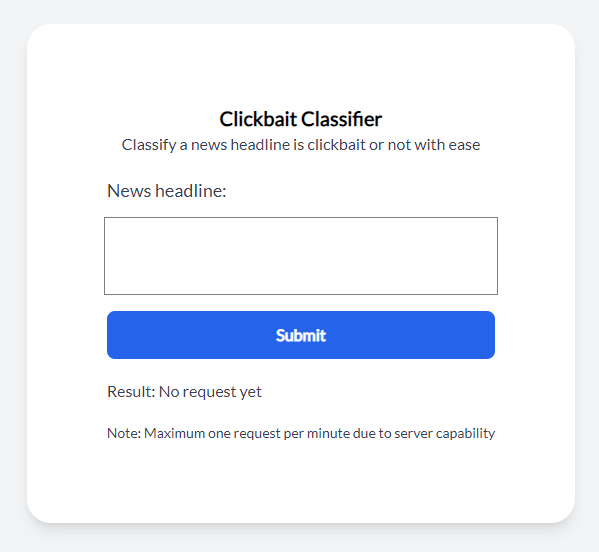}}
\caption{\label{tampilanawal}Application Initial Interface}
\end{figure}

The developed application is accessible through our github page https://ruzcmc.github.io/ClickbaitIndo-textclassifier. Figure \ref{tampilanawal} shows the interface for user input, and request prediction button. 

\begin{figure}[!htb]
  \begin{subfigure}[b]{.5\columnwidth}
    
    \center{\includegraphics[width=\columnwidth]{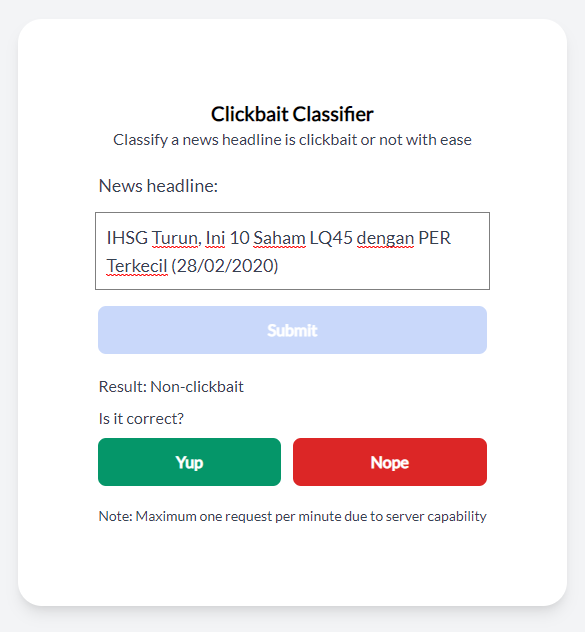}}
    
    \subcaption{\label{tampilanbefore}User Feedback Interface}
    
  \end{subfigure}%
  \begin{subfigure}[b]{.5\columnwidth}
    \center{\includegraphics[width=\columnwidth]{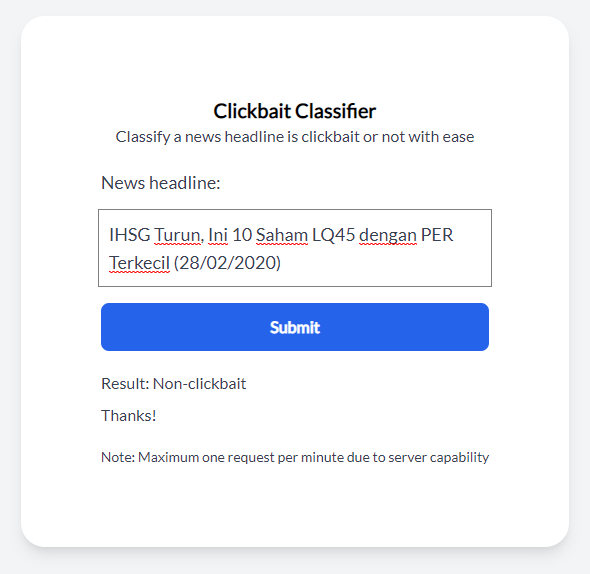}}
    
    \subcaption{After Submitting User Feedback}
    \label{tampilanafter}
  \end{subfigure}
  \caption{After Receiving Prediction}
  \label{tampilanfeedback}
\end{figure}

After the prediction is shown to the user, the interface pop up button for user feedback. The feedback is used to verify the prediction. Figure \ref{tampilanfeedback} shows the interface after user received the prediction and the app request for user feedback.

\section{Discussions}
The web based application for detecting Indonesian clickbait headlines was developed successfully. Although the training performance was slightly waned, the finished product still delivers most of the time.

Future study should aim to evaluate the performance of the app, either from the UI/UX perspective or from the memory efficiency perspective. The prediction model also need to be evaluated and retrained in the future to enhance its capability.

The limitation of the finished product is the lack of security on its endpoints. It should be possible to add extra protection such as Google reCAPTCHA or API-key based authentication. Future study needs to integrate authentication protocol to enhance the endpoint security.

Although the finished product is easy enough to use, sometimes it is a hassle to open new website only for checking clickbaits. Future study can further develop this application into a more integrated solution, such as browser extension capable of flagging clickbaits on-the-fly.

Finally, future researcher should consider to integrate the endpoint into a media monitoring dashboard. A smart dashboard will help stakeholders to monitor their media quality and provide insights to enhance its public relations.

\section*{Additional Resource}
The complete Python notebook, Scripts, Dockerfile, APIs and datasets are stored on https://github.com/ruzcmc/ClickbaitIndo-textclassifier. 
\section*{Acknowledgment}
Training dataset provided by A. William and Y. Sari as cited.


\bibliography{clickbait-textclassify.bib}

\end{document}